\pdfoutput=1
\documentclass[11pt]{article}
\usepackage{naacl2021}

\usepackage{times}
\usepackage{latexsym}
\usepackage[T1]{fontenc}
\usepackage[utf8]{inputenc}
\usepackage{microtype}
\usepackage{graphicx}
\usepackage{hyperref}
\usepackage{amsmath}

\title{Teaching a Massive Open Online Course on Natural Language Processing}

\author{

{\bf Ekaterina Artemova\textsuperscript{1,2}\thanks{Corresponding author, email:  \href{mailto:elartemova@hse.ru}{elartemova@hse.ru}},~Murat Apishev  \textsuperscript{1},~Veronika Sarkisyan  \textsuperscript{1}},\\{\bf Sergey Aksenov\textsuperscript{1},~Denis Kirjanov\textsuperscript{3},~Oleg Serikov \textsuperscript{1,4} } \\
  \textsuperscript{1} HSE University, Moscow, Russia\\
  \textsuperscript{2} Huawei  Noah’s Ark lab, Moscow, Russia\\
  \textsuperscript{3} SberDevices, Sberbank, Moscow, Russia\\
  \textsuperscript{4} DeepPavlov, MIPT, Dolgoprudny, Russia\\

  {\tt \url{https://openedu.ru/course/hse/TEXT/}}
\\}

\begin{document}
\maketitle
\begin{abstract}
This paper presents a new Massive Open Online Course on Natural Language Processing, targeted at non-English speaking students. The course lasts 12 weeks; every week consists of lectures, practical sessions, and quiz assignments. Three weeks out of 12 are followed by Kaggle-style coding assignments. 

Our course intends to serve multiple purposes: (i) familiarize students with the core concepts and methods in NLP, such as language modeling or word or sentence representations, (ii) show that recent advances, including pre-trained Transformer-based models, are built upon these concepts;  (iii)  introduce architectures for most demanded real-life applications, (iv)  develop practical skills to process texts in multiple languages. The course was prepared and recorded during 2020, launched by the end of the year, and in early 2021 has received positive feedback. 
\end{abstract}

\section{Introduction}

The vast majority of recently developed online courses on Artificial Intelligence (AI), Natural Language Processing (NLP) included, are oriented towards English-speaking audiences.  In non-English speaking countries, such courses' audience is unfortunately quite limited, mainly due to the language barrier. Students, who are not fluent in English, find it difficult to cope with language issues and study simultaneously. Thus the students face serious learning difficulties and lack of motivation to complete the online course. While creating new online courses in languages other than English seems redundant and unprofitable, there are multiple reasons to support it. First, students may find it easier to comprehend new concepts and problems in their native language. Secondly, it may be easier to build a strong online learning community if students can express themselves fluently. Finally, and more specifically to NLP, an NLP course aimed at building practical skills should include language-specific tools and applications. Knowing how to use tools for English is essential to understand the core principles of the NLP pipeline. However, it is of little use if the students work on real-life applications in the non-English industry. 

In this paper, we present an overview of an online course aimed at Russian-speaking students. This course was developed and run for the first time in 2020, achieving positive feedback. Our course is a part of the HSE university's online specialization on AI and is built upon previous courses in the specialization, which introduced core concepts in calculus, probability theory, and programming in Python. Outside of the specialization, the course can be used for additional training of students majoring in computer science or software engineering and others who fulfill prerequisites.

The main contributions of this paper are:
\begin{itemize}
    \item We present the syllabus of a recent wide-scope massive open online course on NLP, aimed at a broad audience;  
    \item We describe methodological choices made for teaching NLP to non-English speaking students;
    \item In this course, we combine recent deep learning trends with other best practices, such as topic modeling. 
    
\end{itemize}

The remainder of the paper is organized as follows: Section~\ref{sec:overview} introduces methodological choices made for the course design. Section~\ref{sec:syllabus} presents the course structure and topics in more details. Section~\ref{sec:hw} lists home works. Section~\ref{sec:platform} describes the hosting platform and its functionality.

\section{Course overview} \label{sec:overview}

The course presented in this paper is split into two main parts, six weeks each, which cover (i) core NLP concepts and approaches and (ii) main applications and more sophisticated problem formulations. The first six weeks' main goal is to present different word and sentence representation methods, starting from bag-of-words and moving to word and sentence embeddings, reaching contextualized word embeddings and pre-trained language models. Simultaneously we introduce basic problem definitions: text classification, sequence labeling, and sequence-to-sequence transformation. The first part of the course roughly follows Yoav Goldberg's textbook~\cite{goldberg2017neural}, albeit we extend it with pre-training approaches and recent Transformer-based architectures. 

The second part of the course introduces  BERT-based models and such NLP applications as question
answering, text summarization, and information extraction. This part adopts some of the explanations from the recent draft of ``Speech and Language Processing'' \cite{jurafsky2000speech}. An entire week is devoted to topic modeling, and BigARTM \cite{vorontsov2015bigartm}, a tool for topic modeling developed in MIPT, one of the top Russian universities and widely used in real-life applications. Overall practical sessions are aimed at developing text processing skills and practical coding skills. 

Every week comprises both a lecture and a practical session. Lectures have a ``talking head'' format, so slides and pre-recorded demos are presented, while practical sessions are real-time coding sessions. The instructor writes code snippets in Jupyter notebooks and explains them at the same time. Overall every week, there are 3-5 lecture videos and 2-3 practical session videos.  Weeks \hyperref[weekclf]{3}, \hyperref[weeklm]{5}, \hyperref[weektm]{9} are extended with coding assignments. 

 Weeks \hyperref[sesame2]{7} and \hyperref[weektm]{9} are followed by interviews. In these interviews, one of the instructors' talks to the leading specialist in the area. Tatyana Shavrina, one of the guests interviewed, leads an R\&D team in Sber, one of the leading IT companies. The second guest, Konstantin Vorontsov, is a professor from one of the top universities. The guests are asked about their current projects and interests, career paths, what keeps them inspired and motivated, and what kind of advice they can give. 

The final mark is calculated according to the formula: 

\begin{equation*}
\begin{aligned}
    \textit{\#\ of\ accepted\ coding\ assignment} \\
    + 0.7 \texttt{mean}\textit{(quiz\ assignment\ mark)}
\end{aligned}
\end{equation*} 

Coding assignments are evaluated on the binary scale (accepted or rejected), and quiz assignments are evaluated on the 10 point scale. To earn a certificate, the student has to earn at least 4 points.  

In practical sessions, we made a special effort to introduce tools developed for processing texts in Russian. The vast majority of examples, utilized in lectures, problems, attempted during practical sessions, and coding assignments, utilized datasets in Russian. The same choice was made by Pavel Braslavski, who was the first to create an NLP course in Russian in 2017~\cite{pbras}. We utilized datasets in English only if Russian lacks the non-commercial and freely available datasets for the same task of high quality. 

Some topics are intentionally not covered in the course. We focus on written texts and do not approach the tasks of text-to-speech and speech-to-text transformations. Low-resource languages spoken in Russia are out of the scope, too. Besides, we almost left out potentially controversial topics, such as AI ethics and green AI problems. Although we briefly touch upon potential biases in pre-trained language models, we have to leave out a large body of research in the area, mainly oriented towards the English language and the US or European social problems. Besides, little has been explored in how neural models are affected by those biases and problems in Russia. 

The team of instructors includes specialists from different backgrounds in computer science and theoretical linguists. Three instructors worked on lectures, two instructors taught practical sessions, and three teaching assistants prepared home assignments and conducted question-answering sessions in the course forum.

\section{Syllabus}\label{sec:syllabus}

\noindent \paragraph{Week 1. Introduction.} The first introductory \textbf{lecture} consists of two parts. The first part overviews the core tasks and problems in NLP, presents the main industrial applications, such as search engines, Business Intelligence tools, and conversational engines, and draws a comparison between broad-defined linguistics and NLP. To conclude this part, we touch upon recent trends, which can be grasped easily without the need to go deep into details, such as multi-modal applications ~\cite{zhou2020unified}, cross-lingual methods ~\cite{feng2020language,conneau2020unsupervised} and computational humor ~\cite{braslavski2018evaluate,west2019reverse}. Throughout this part lecture, we try to show NLP systems' duality: those aimed at understanding language (or speech) and those aimed at generating language (or speech). The most complex systems used for machine translation, for example, aim at both. 
The second part of the lecture introduces such basic concepts as bag-of-words, count-based document vector representation, \texttt{tf-idf} weighting. Finally, we explore bigram association measures, \texttt{PMI} and \texttt{t-score}. We point out that these techniques can be used to conduct an exploratory analysis of a given collection of texts and prepare input for machine learning methods.

\textbf{Practical session} gives an overview of text prepossessing techniques and simple count-based text representation models. We emphasize how prepossessing pipelines can differ for languages such as English and Russian (for example, what is preferable, stemming or lemmatization) and give examples of Python frameworks that are designed to work with the Russian language (pymystem3 ~\cite{segalovich2003fast}, pymorphy2 ~\cite{korobov2015morphological}). We also included an intro to regular expressions because we find this knowledge instrumental both within and outside NLP tasks.

During the first weeks, most participants are highly motivated, we can afford to give them more practical material, but we still need to end up with some close-to-life clear examples. We use a simple sentiment analysis task on Twitter data to demonstrate that even the first week's knowledge (together with understanding basic machine learning) allows participants to solve real-world problems. At the same time, we illustrate how particular steps of text prepossessing can have a crucial impact on the model's outcome.

\noindent \paragraph{Week 2. Word embeddings.}
{\bf The lecture} introduces the concepts of distributional semantics and word vector representations. We familiarize the students with early models, which utilized singular value decomposition (SVD) and move towards more advanced word embedding models, such as \texttt{word2vec}~\cite{mikolov2013distributed} and \texttt{fasttext}~\cite{bojanowski2017enriching}. We briefly touch upon the hierarchical softmax and the hashing trick and draw attention to negative sampling techniques. We show ways to compute word distance, including Euclidean and cosine similarity measures. 

\begin{figure}
\centering
\includegraphics[width = .4\textwidth]{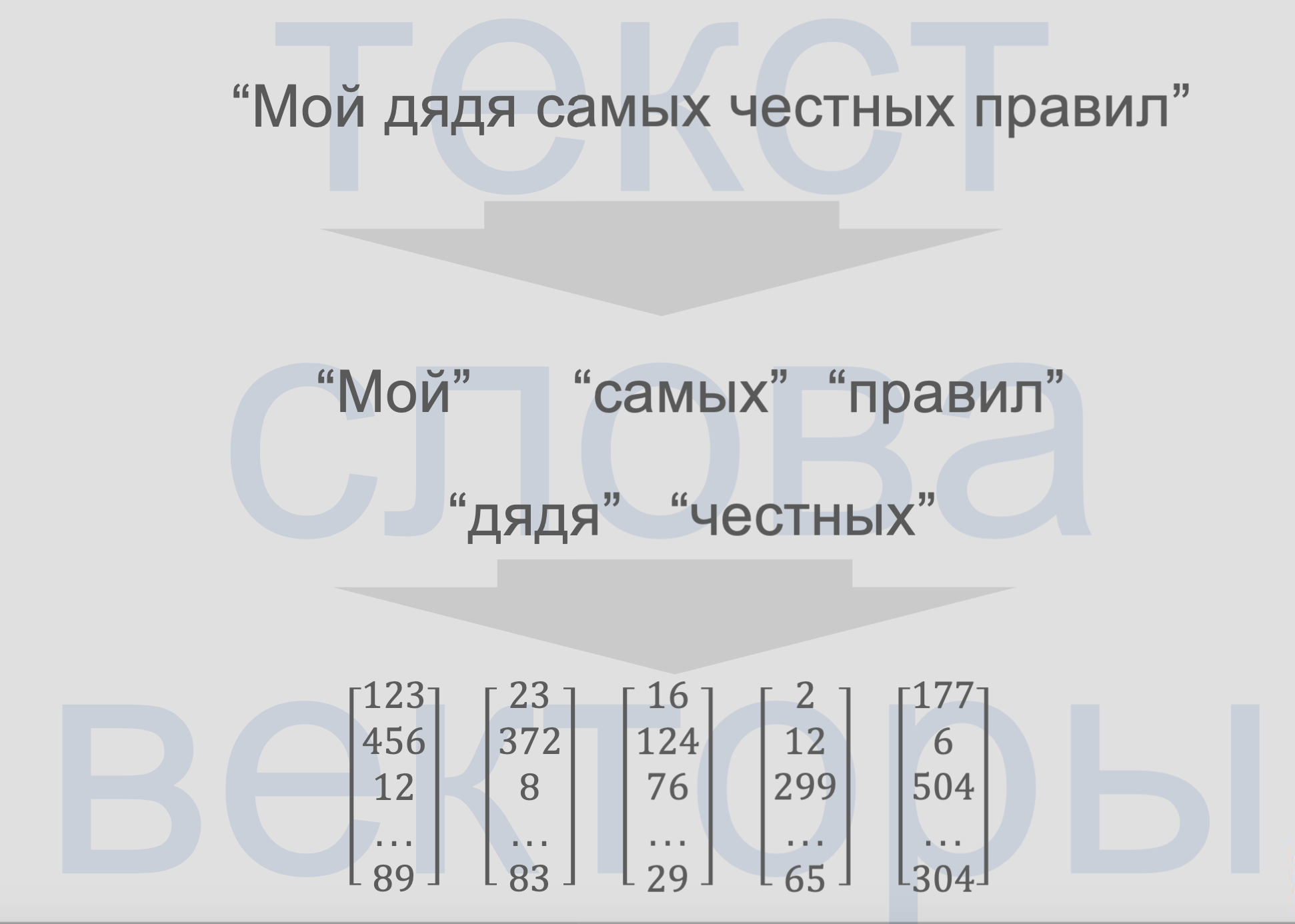}
\caption{One slide from Lecture 2. Difference between raw texts (top line), bag-of-words (middle line), and bag-of-vectors (bottom line). Background words: text, words, vectors. }
\label{fig:bow}
\end{figure}

We discuss the difference between \texttt{word2vec} and \texttt{GloVe}~\cite{pennington2014glove} models and emphasize main issues, such as dealing with out-of-vocabulary (OOV) words and disregarding rich morphology. \texttt{fasttext} is then claimed to address these issues. To conclude, we present approaches for intrinsic and extrinsic evaluation of word embeddings. Fig.~\ref{fig:bow} explains the difference between bag-of-words and bag-of-vectors. 

In \textbf{practical session} we explore only advanced word embedding models (\texttt{word2vec}, \texttt{fasttext} and \texttt{GloVe}) and we cover three most common scenarios for working with such models: using pre-trained models, training models from scratch and tuning pre-trained models. Giving a few examples, we show that \texttt{fasttext} as a character-level model serves as a better word representation model for Russian and copes better with Russian rich morphology. We also demonstrate some approaches of intrinsic evaluation of models' quality, such as solving analogy tasks (like well known ``king - man + woman = queen'') and evaluating semantic similarity and some useful techniques for visualization of word embeddings space.

This topic can be fascinating for students when supplemented with illustrative examples. Exploring visualization of words clusters on plots or solving analogies is a memorable part of the ``classic'' NLP part of most students' course.  

\noindent \paragraph{Week 3. Text classification.}\label{weekclf} The  {\bf lecture} considers core concepts for supervised learning. We begin by providing examples for text classification applications, such as sentiment classification and spam filtering. Multiple problem statements, such as binary, multi-class, and multi-label classification, are stated. To introduce ML algorithms, we start with logistic regression and move towards neural methods for text classification. To this end, we introduce \texttt{fasttext} as an easy, out-of-the-box solution. We introduce the concept of sentence (paragraph) embedding by presenting \texttt{doc2vec} model~\cite{le2014distributed} and show how such embeddings can be used as input to the classification model. 
Next, we move towards more sophisticated techniques, including convolutional models for sentence classification ~\cite{DBLP:journals/corr/Kim14f}. We do not discuss backpropagation algorithms but refer to the DL course of the specialization to refresh understanding of neural network training. We show ways to collect annotated data on crowdsourcing platforms and speed up the process using active learning ~\cite{esuli2009active}.  
Finally, we conclude with  text augmentation techniques, including SMOTE~\cite{chawla2002smote} and EDA ~\cite{wei2019eda}.

In the \textbf{practical session} we continue working with the text classification on the IMDb movies reviews dataset. 
We demonstrate several approaches to create classification models with different word embeddings. 
We compare two different ways to get sentence embedding, based on any word embedding model: by averaging word vectors and using \texttt{tf-idf} weights for a linear combination of word vectors. 
We showcase \texttt{fasttext} tool for text classification using its built-in classification algorithm.

Additionally, we consider use \texttt{GloVe} word embedding model to build a simple Convolutional Neural Network for text classification. In this week and all of the following, we use PyTorch \footnote{\url{https://pytorch.org/}} as a framework for deep learning.

\noindent \paragraph{Week 4. Language modeling.}\label{weeklm} The {\bf lecture} focuses on the concept of language modelling. We start with early count-based models ~\cite{song1999general} and create a link to Markov chains. We refer to the problem of OOV words and show the \texttt{add-one} smoothing method, avoiding more sophisticated techniques, such as Knesser-Ney smoothing ~\cite{kneser1995improved}, for the sake of time. Next, we introduce neural language models. To this end, we first approach Bengio's language model~\cite{bengio2003neural}, which utilizes fully connected layers. Second, we present recurrent neural networks and show how they can be used for language modeling. Again, we remind the students of backpropagation through time and gradient vanishing or explosion, introduced earlier in the DL course.  We claim, that LSTM ~\cite{hochreiter1997long} and GRU ~\cite{chung2014empirical} cope with these problems. As a brief revision of the LSTM architecture is necessary, we utilize Christopher Olah's tutorial~\cite{olah2015understanding}. We pay extra attention to the inner working of the LSTM, following Andrej Karpathy's tutorial ~\cite{karpathy2015unreasonable}. To add some research flavor to the lecture, we talk about text generation~\cite{sutskever2011generating}, its application, and different decoding strategies ~\cite{holtzman2019curious}, including beam search and nucleus sampling.  Lastly, we introduce the sequence labeling task ~\cite{ma2016end} for part-of-speech (POS) tagging and named entity recognition (NER) and show how RNN's can be utilized as sequence models for the tasks. 

The \textbf{practical session} in this week is divided into two parts. The first part is dedicated to language models for text generation. We experiment with count-based probabilistic models and RNN's to generate dinosaur names and get familiar with perplexity calculation (the task and the data were introduced in Sequence Models course from DeepLearning.AI \footnote{\url{https://www.coursera.org/learn/nlp-sequence-models}}). To bring things together, students are asked to make minor changes in the code and run it to answer some questions in the week's quiz assignment.

The second part of the session demonstrates the application of RNN's to named entity recognition. We first introduce the BIO and BIOES annotation schemes and show frameworks with pre-trained NER models for English (Spacy \footnote{\url{https://spacy.io}}) and Russian (Natasha\footnote{\url{https://natasha.github.io}}) languages. Further, we move on to CNN-biLSTM-CRF architecture described in the lecture and test it on CoNLL 2003 shared task data~\cite{sang2003introduction}.

\noindent \paragraph{Week 5. Machine Translation.} This {\bf lecture} starts with referring to the common experience of using machine translation tools and a historical overview of the area. Next, the idea of encoder-decoder (seq2seq) architecture opens the technical part of the lecture. We start with RNN-based seq2seq models ~\cite{sutskever2014sequence} and introduce the concept of attention~\cite{bahdanau2015neural}. We show how attention maps can be used for ``black box'' interpretation. Next, we reveal the core architecture of modern NLP, namely, the Transformer model ~\cite{vaswani2017attention} and ask the students explicitly to take this part seriously. Following Jay Allamar's tutorial ~\cite{illustrated}, we decompose the transformer architecture and go through it step by step. In the last part of the lecture, we return to machine translation and introduce quality measures, such as WER and BLEU ~\cite{papineni2002bleu}, touch upon human evaluation and the fact that BLEU correlates well with human judgments. Finally, we discuss briefly more advanced techniques, such as non-autoregressive models~\cite{gu2017non} and back translation ~\cite{hoang2018iterative}. Although we do not expect the student to comprehend these techniques immediately, we want to broaden their horizons so that they can think out of the box of supervised learning and autoregressive decoding. 

In the first part of \textbf{practical session} we solve the following task: given a date in an arbitrary format transform it to the standard format ``dd-mm-yyyy'' (for example, ``18 Feb 2018'', ``18.02.2018'', ``18/02/2018'' $\rightarrow$ ``18-02-2018''). We adopt the code from PyTorch machine translation tutorial \footnote{\url{https://pytorch.org/tutorials/intermediate/seq2seq_translation_tutorial.html}} to our task: we use the same RNN encoder, RNN decoder, and its modification - RNN encoder with attention mechanism - and compare the quality of two decoders. We also demonstrate how to visualize attention weights.

The second part is dedicated to the Transformer model and is based on the Harvard NLP tutorial ~\cite{transfomer_tutorial} that decomposes the article ``Attention is All You Need'' ~\cite{vaswani2017attention}. Step by step, like in the lecture, we go through the Transformer code, trying to draw parallels with a simple encoder-decoder model we have seen in the first part. We describe and comment on every layer and pay special attention to implementing the attention layer and masking and the shapes of embeddings and layers.

\noindent \paragraph{Week 6. Sesame Street I.} The sixth {\bf lecture} and the next one are the most intense in the course. The paradigm of pre-trained language models is introduced in these two weeks. The first model to discuss in detail is ELMo~\cite{peters2018deep}. Next, we move to BERT ~\cite{devlin2019bert} and introduce the masked language modeling and next sentence prediction objectives. While presenting BERT, we briefly revise the inner working of Transformer blocks. We showcase three scenarios to fine-tune BERT: (i) text classification by using different pooling strategies (\texttt{[CLS]}, \texttt{max} or \texttt{mean}), (ii) sentence pair classification for paraphrase identification and for natural language inference, (iii) named entity recognition. SQuAD-style question-answering, at which BERT is aimed too, as avoided here, as we will have another week for QA systems. Next, we move towards GPT-2 ~\cite{radfordlanguage} and elaborate on how high-quality text generation can be potentially harmful. To make the difference between BERT's and GPT-2's objective more clear, we draw parallels with the Transformer architecture for machine translation and show that BERT is an encoder-style model, while GPT-2 is a decoder-style model. We show Allen NLP ~\cite{gardner2018allennlp} demos of how GPT-2 generates texts and how attention scores implicitly resolve coreference. 

\begin{figure}
\centering
\includegraphics[width = .2\textwidth]{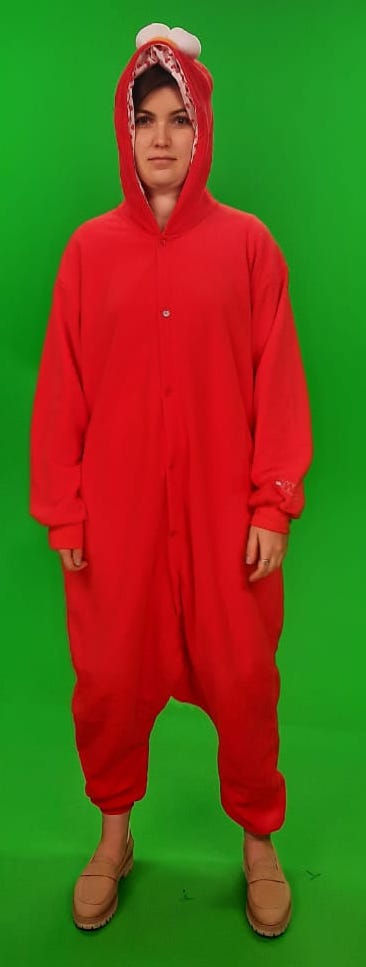}
\caption{To spice up the lectures, the lecturer is dressed in an ELMo costume}
\label{fig:elmo}
\end{figure}

In this week, we massively rely on Jay Allamar's ~\cite{illustrated} tutorial and adopt some of these brilliant illustrations. One of the main problems, though, rising in this week is the lack of Russian terminology, as the Russian-speaking community has not agreed on the proper ways to translate such terms as ``contextualized encoder'' or ``fine-tuning''. To spice up this week, we were dressed in Sesame Street kigurumis (see Fig.~\ref{fig:elmo}).

The main idea of the \textbf{practical session} is to demonstrate ELMo and BERT models, considered earlier in the lecture.
The session is divided into two parts, and in both parts, we consider text classification, using ELMo and BERT models, respectively.

In the first part, we demonstrate how to use ELMo word embeddings for text classification on the IMBdb dataset used in previous sessions. We use pre-trained ELMo embeddings by AllenNLP~\cite{gardner2018allennlp} library and implement a simple recurrent neural network with a GRU layer on top for text classification. In the end, we compare the performance of this model with the scores we got in previous sessions on the same dataset and demonstrate that using ELMo embeddings can improve model performance.

The second part of the session is focused on models based on Transformer architecture. We use huggingface-transformers library~\cite{wolf-etal-2020-transformers} and a pre-trained BERT model to build a classification algorithm for Google play applications reviews written in English. We implement an entire pipeline of data preparation, using a pre-trained model and demonstrating how to fine-tune the downstream task model. Besides, we implement a wrapper for the BERT classification model to get the prediction on new text.

\noindent \paragraph{Week 7. Sesame Street II.}\label{sesame2}  To continue diving into the pre-trained language model paradigm, the {\bf lecture} first questions, how to evaluate the model. We discuss some methods to interpret the BERT's inner workings, sometimes referred to as BERTology ~\cite{rogers2021primer}. We introduce a few common ideas: BERT's lower layers account for surface features, lower to middle layers are responsible for morphology, while the upper-middle layers have better syntax representation ~\cite{conneau2018senteval}. We talk about ethical issues ~\cite{may2019measuring}, caused by pre-training on raw web texts. We move towards the extrinsic evaluation of pre-trained models and familiarize the students with GLUE-style evaluations ~\cite{wang2019glue,wang2019superglue}. The next part of the lecture covers different improvements of BERT-like models. We show how different design choices may affect the model's performance in different tasks and present RoBERTa ~\cite{liu2019roberta}, and ALBERT ~\cite{lan2019albert} as members of a BERT-based family. We touch upon the computational inefficiency of pre-trained models and introduce lighter models, including DistillBERT~\cite{sanh2019distilbert}. To be solid, we touch upon other techniques to compress pre-trained models, including pruning~\cite{sajjad2020poor} and quantization~\cite{zafrir2019q8bert}, but do not expect the students to be able to implement these techniques immediately.  We present the concept of language transferring and introduce multilingual Transformers, such as XLM-R~\cite{conneau2020unsupervised}. Language transfer becomes more and more crucial for non-English applications, and thus we draw more attention to it.
Finally, we cover some of the basic multi-modal models aimed at image captioning and visual question answering, such as the unified Vision-Language Pre-training (VLP) model ~\cite{zhou2020unified}.

In the \textbf{practical session} we continue discussing BERT-based models, shown in the lectures. The session's main idea is to consider different tasks that may be solved by BERT-based models and to demonstrate different tools and approaches for solving them. So the practical session is divided into two parts. 
The first part is devoted to named entity recognition. We consider a pre-trained cross-lingual BERT-based NER model from the DeepPavlov library~\cite{burtsev2018deeppavlov} and demonstrate how it can be used to extract named entities from Russian and English text.
The second part is focused on multilingual zero-shot classification. We consider the pre-trained XLM-based model by HuggingFace, discuss the approach's key ideas, and demonstrate how the model works, classifying short texts in English, Russian, Spanish, and French.

\noindent \paragraph{Week 8. Syntax parsing.} The {\bf lecture} is devoted to computational approaches to syntactic parsing and is structured as follows.
After a brief introduction about the matter and its possible applications (both as an auxiliary task and an independent one), we consider syntactic frameworks developed in linguistics: dependency grammar~\cite{tesniere2015elements} and constituency grammar \cite{bloomfield1936language}. Then we discuss only algorithms that deal with dependency parsing (mainly because there are no constituency parsers for Russian), so we turn to graph-based \cite{mcdonald2005online} and transition-based \cite{aho1972theory} dependency parsers and consider their logics, structure, sorts, advantages, and drawbacks. Afterward, we familiarize students with the practical side of parsing, so we introduce syntactically annotated corpora, Universal Dependencies project \cite{nivre2016universal} and some parsers which perform for Russian well (UDPipe~\cite{straka2017tokenizing}, DeepPavlov Project ~\cite{burtsev2018deeppavlov}).
The last part of our lecture represents a brief overview of the problems which were not covered in previous parts: BERTology, some issues of web-texts parsing, latest advances in computational syntax (like enhanced dependencies~\cite{schuster2016enhanced}).

The \textbf{practical session} starts with a quick overview of CoNLL-U annotation format \cite{nivre-etal-2016-universal}: we show how to load, parse and visualize such data on the example from the SynTagRus corpus \footnote{\url{https://universaldependencies.org/treebanks/ru_syntagrus/index.html}}. Next, we learn to parse data with pre-trained UDPipe models \cite{straka-etal-2016-udpipe} and Russian-language framework Natasha. To demonstrate some practical usage of syntax parsing, we first understand how to extract subject-verb-object (SVO) triples and then design a simple template-based text summarization model.

\noindent \paragraph{Week 9. Topic modelling}\label{weektm} The focus of this {\bf lecture} is topic modeling. First, we formulate the topic modeling problem and ways it can be used to cluster texts or extract topics. We explain the basic probabilistic latent semantic analysis (PLSA) model~\cite{hofmann1999probabilistic}, that modifies early approaches, which were based on SVD ~\cite{dumais2004latent}. We approach the PLSA problem using the Expectation-Minimization (EM) algorithm and introduce the basic performance metrics, such as perplexity and topic coherence.

\begin{figure}
\centering
\includegraphics[width = .4\textwidth]{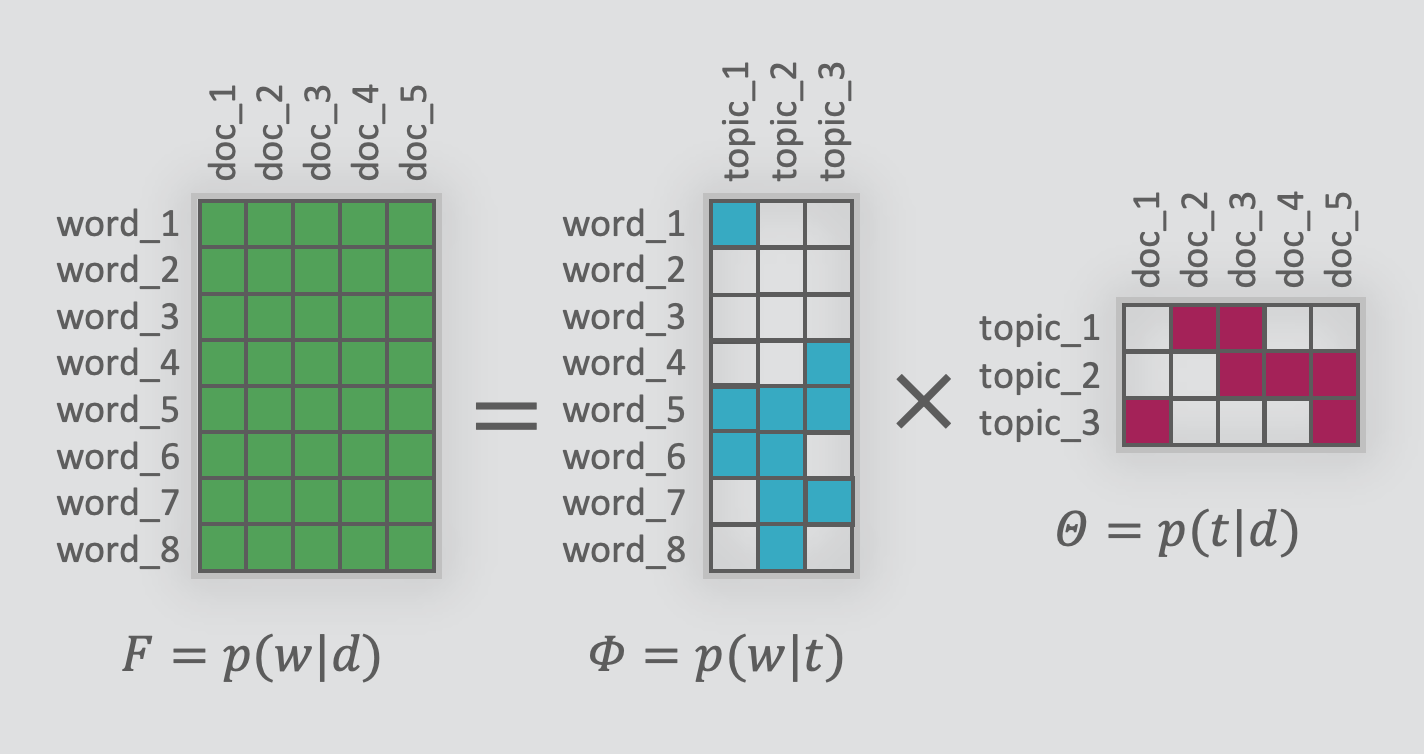}
\caption{One slide from Lecture 9. Sparsification of an ARTM model explained.}
\label{fig:sparse}
\end{figure}

As the PLSA problem is ill-posed, we familiarize students with regularization techniques using Additive Regularization for Topic Modeling (ARTM) model ~\cite{vorontsov2015additive} as an example. We describe the general EM algorithm for ARTM and some basic regularizers. Then we move towards the Latent Dirichlet Allocation (LDA) model ~\cite{blei2003latent} and show that the maximum a posteriori estimation for LDA  is the special case of the ARTM model with a smoothing or sparsing regularizer (see Fig.~\ref{fig:sparse} for the explanation snippet). We conclude the lecture with a brief introduction to multi-modal ARTM models and show how to generalize different Bayesian topic models based on LDA. We showcase classification, word translation, and trend detection tasks as multi-modal models.

In \textbf{practical session} we consider the models discussed in the lecture in a slightly different order. First, we take a closer look at Gensim realization of the LDA model \cite{rehurek_lrec}, pick up the model's optimal parameters in terms of perplexity and topic coherence, and visualize the model with pyLDAvis library. Next, we explore BigARTM ~\cite{vorontsov2015bigartm} library, particularly LDA, PLSA, and multi-modal models, and the impact of different regularizers. For all experiments, we use a corpus of Russian-language news from Lenta.ru \footnote{\url{https://github.com/yutkin/Lenta.Ru-News-Dataset}} which allows us to compare the models to each other.  

\noindent \paragraph{Week 10.} In this {\bf lecture} we discussed mono-lingual seq2seq problems, text summarization and sentence simplification. We start with extractive summarization techniques. The first approach introduced is TextRank~\cite{mihalcea2004textrank}. We present each step of this approach and explain that any sentence or keyword embeddings can be used to construct a text graph, as required by the method. Thus we refer the students back to earlier lectures, where sentence embeddings were discussed. Next, we move to abstractive summarization techniques. To this end, we present performance metrics, such as ROUGE~\cite{lin2004rouge} and METEOR~\cite{banerjee2005meteor} and  briefly overview pre-Transformer architectures, including Pointer networks \cite{see2017get}. Next, we show recent pre-trained Transformer-based models, which aim at multi-task learning, including summarization. To this end, we discuss pre-training approaches of T5~\cite{raffel2020exploring} and BART~\cite{lewis2020bart}, and how they help to improve the performance of mono-lingual se2seq tasks. Unfortunately, when this lecture was created, multilingual versions of these models were not available, so they are left out of the scope. 
Finally, we talk about sentence simplification task \cite{coster2011simple,alva2020data} and its social impact. We present SARI~\cite{xu2016optimizing} as a metric for sentence simplification performance and state, explain how T5 or BART can be utilized for the task.

The \textbf{practical session} is devoted to extractive summarization and TextRank algorithm. We are urged to stick to extractive summarization, as Russian lacks annotated datasets, but, at the same time, the task is demanded by in industry—extractive summarization compromises than between the need for summarization techniques and the absence of training datasets. Nevertheless, we used annotated English datasets to show how performance metrics can be used for the task.
The CNN/DailyMail articles are used as an example of a dataset for the summarization task. As there is no standard benchmark for text summarization in Russian, we have to use English to measure different models' performance. We implement the TextRank algorithm and compare it with the algorithm from the NetworkX library \cite{hagberg2008exploring}. Also, we demonstrate how to estimate the performance of the summarization by calculating the ROUGE metric for the resulting algorithm using the PyRouge library\footnote{url{https://github.com/andersjo/pyrouge}}.
This practical session allows us to refer back the students to sentence embedding models and showcase another application of sentence vectors. 

\noindent \paragraph{Week 11.} The penultimate {\bf lecture} approaches  Question-Answering (QA) systems and chat-bot technologies. We present multiple real-life industrial applications, where chat-bots and QA technologies are used, ranging from simple task-oriented chat-bots for food ordering to help desk or hotline automation. 
Next, we formulate the core problems of task-oriented chat-bots, which are intent classification and slot-filling~\cite{liu2016attention} and revise methods, to approach them. After that, we introduce the concept of a dialog scenario graph and show how such a graph can guide users to complete their requests. Without going deep into technical details, we show how ready-made solutions, such as Google Dialogflow\footnote{\url{https://cloud.google.com/dialogflow}}, can be used to create task-oriented chat-bots. Next, we move towards QA models, of which we pay more attention to information retrieval-based (IR-based) approaches and SQuAD-style~\cite{rajpurkar2016squad} approaches. Since natural language generation models are not mature enough (at least for Russian) to be used in free dialog, we explain how IR-based techniques imitate a conversation with a user. 
Finally, we show how BERT can be used to tackle the SQuAD problem. The lecture is concluded by comparing industrial dialog assistants created by Russian companies, such as Yandex.Alisa or Mail.ru Marusya.

In the \textbf{practical session} we demonstrate several examples of using Transformer-based models for QA task. Firstly, we try to finetune Electra model \cite{clark2020electra} on COVID-19 questions dataset \footnote{\url{https://github.com/xhlulu/covid-qa}} and BERT on SQuAD 2.0 \cite{DBLP:journals/corr/squad2} (we use code from hugginface tutorial \footnote{\url{https://huggingface.co/transformers/custom_datasets.html\#question-answering-with-squad-2-0}} for the latter). Next, we show an example of usage of pretrained model for Russian-language data from DeepPavlov project. Finally, we explore how to use BERT for joint intent classification and slot filling task \cite{DBLP:journals/corr/jointbert}.

\noindent \paragraph{Week 12.} The last {\bf lecture} wraps up the course by discussing knowledge graphs (KG) and some of their applications for QA systems. We revise core information extraction problems, such as NER and relation detection, and show how they can be used to extract a knowledge graph from unstructured texts~\cite{paulheim2017knowledge}. We touch upon the entity linking problem but do not go deep into details. To propose to students an alternative view to information extraction, we present machine reading comprehension approaches for NER~\cite{li2019unified} and relation detection~\cite{li2019entity}, referring to the previous lecture. Finally, we close the course by revising all topics covered. We recite the evolution of text representation models from bag-of-words to BERT. We show that all the problems discussed throughout the course fall into one of three categories: (i) text classification or sentence pair classification, (ii) sequence tagging, (iii) sequence-to-sequence transformation. We draw attention to the fact that the most recent models can tackle all of the problem categories. Last but not least we revise, how all of these problem statements are utilized in real-life applications.  

The \textbf{practical session} in this week is dedicated to information extraction tasks with Stanford CoreNLP library \cite{manning2014stanford}. The session's main idea is to demonstrate using the tool for constructing knowledge graphs based on natural text. We consider different ways of using the library and experimented with using the library to solve different NLP tasks that were already considered in the course: tokenization, lemmatization, POS-tagging, and dependency parsing. The library includes models for 53 languages, so we consider examples of solving these tasks for  English and Russian texts. Besides, relation extraction is considered using the Open Information Extraction (OpenIE) module from the CoreNLP library.

\section{Home works}\label{sec:hw}
The course consists of multiple ungraded quiz assignments, 11 graded quiz assignments, three graded coding assignments. Grading is performed automatically in a Kaggle-like fashion.

\subsection{Quiz Assignments}
Every video lecture is followed by an ungraded quiz, consisting of 1-2 questions. A typical question address the core concepts introduced:
\begin{itemize}
    \item What kind of vectors are more common for word embedding models?  \\
    A1: dense (true), A2: sparse (false)
    \item What kind of layers are essential for  GPT-2 model? 
    \\
    A1: transformer stacks (true), A2: recurrent layers (false), A3: convolutional layers (false), A4: dense layers (false)
\end{itemize}

A graded test is conducted every week, except the very last one. It consists of 12-15 questions, which we tried to split into three parts, being more or less of the same complexity. First part questions about main concepts and ideas introduced during the week. These questions are a bit more complicated than after video ones:
\begin{itemize}
    \item What part of an encoder-decoder model solves the language modeling problem, i.e., the next word prediction?
    \\
    A1: encoder (false), A2: decoder (true)
    \item What are the BPE algorithm units?
    \\
    A1: syllables (false), A2: morphemes (false), A3: $n-$grams (true), A4: words (false)
\end{itemize}
Second part of the quiz asks the students to conduct simple computations by hand:
\begin{itemize}
    \item Given a detailed description of an neural architecture, compute the number of parameters;
    \item Given a gold-standard NER annotation and a system output, compute token-based and span-based micro $F_1$.
\end{itemize}

The third part of the quiz asks to complete a simple programming assignment or asks about the code presented in practical sessions:
\begin{itemize}
    \item Given a pre-trained language model, compute perplexity of a test sentence
    \item Does DeepPavlov cross-lingual NER model require to announce the language of the input text?
\end{itemize}

For convenience and to avoid format ambiguity, all questions are in multiple-choice format. For questions, which require a numerical answer, we provided answer options in the form of intervals, with one of the endpoints excluded. 

Each quiz is estimated on a 10 point scale. All questions have equal weights. 

The final week is followed by a comprehensive quiz covering all topics studied. This quiz is obligatory for those students who desire to earn a certificate.  

\subsection{Coding assignments}
There are three coding assignments  concerning the following topics:
(i) text classification, (ii) sequence labeling, (iii) topic modeling.
Assignments grading is binary.
Text classification and sequence labeling assignments require students to beat the score of the provided baseline submission. 
Topic modeling assignment is evaluated differently.

All the coding tasks provide students with the starter code and sample submission bundles. 
The number of student's submissions is limited. 
Sample submission bundles illustrate the required submission format and could serve as the random baseline for each task.
Submissions are evaluated using the Moodle\footnote{\url{https://moodle.org/}} \cite{moodle} CodeRunner\footnote{\url{https://coderunner.org.nz/}} \cite{lobb2016coderunner} plugin. 

\subsubsection{Text classification and sequence labeling coding assignments}
\paragraph{Text classification assignment} is based on the Harry Potter and the Action Prediction Challenge from Natural Language dataset \cite{vilares-gomez-rodriguez-2019-harry}, which uses fiction fantasy texts.  Here, the task is the following: given some text preceding a spell occurrence in the text, predict this spell name. Students are provided with starter code in Jupyter notebooks \cite{jupyter}.  Starter code implements all the needed data pre-processing, shows how to implement the baseline Logistic Regression model, and provides code needed to generate the submission.

Students' goal is to build three different models performing better than the baseline.  The first one should differ from the baseline model by only hyperparameter values. The second one should be a Gradient Boosting model. The third model to build is a CNN model. All the three models' predictions on the provided testing dataset should be then submitted to the scoring system.  Submissions, where all the models beat the baseline models classification F1-score, are graded positively.

\paragraph{Sequence labeling}
Sequence labeling assignment is based on the LitBank data \cite{bamman2019annotated}.
Here, the task is to given fiction texts, perform a NER labeling. Students are provided with a starter code for data pre-processing and submission packaging. Starter code also illustrates building a recurrent neural model using the PyTorch framework, showing how to compose a single-layer unidirectional RNN model.

Students' goal is to build a bidirectional LSTM model that would outperform the baseline. Submissions are based on the held-out testing subset provided by the course team.

\subsubsection{Topic modeling assignment}
Topic modeling assignment motivation is to give students practical experience with LDA \cite{blei2003latent} algorithm. 
The assignment is organized as follows: first, students have to download and pre-process Wikipedia texts.

Then, the following experiment should be conducted. The experiment consists of training and exploring an LDA model for the given collection of texts.  The task is to build several LDA models for the given data: models differ only in the configured number of topics. Students are asked to explore the obtained models using the pyLDAvis \cite{sievert2014ldavis} tool. This stage is not evaluated. Finally, students are asked to submit the topic labels that LDA models assign to words provided by the course team.  Such a prediction should be performed for each of the obtained models.

\section{Platform description} \label{sec:platform}
The course is hosted on OpenEdu \footnote{\url{https://npoed.ru/}} - an educational platform created by the Association ``National Platform for Open Education'', established by leading Russian universities. Our course and all courses on the platform are available free of charge so that everyone can access all materials (including videos, practical Jupyter notebooks, tests, and coding assessments). The platform also provides a forum where course participants can ask questions or discuss the material with each other and lecturers.


\section{Expected outcomes}
First of all, we expect the students to understand basic formulations of the NLP tasks, such as text classification, sentence pair modeling, sequence tagging, and sequence-to-sequence transformation. We expect the students to be able to recall core terminology and use it fluently. In some weeks, we provide links to extra materials, mainly in English, so that the students can learn more about the topic themselves. We hope that after completing the course, the students become able to read those materials.  Secondly, we anticipate that after completing the course, the students are comfortable using popular Python tools to process texts in Russian and English and utilize pre-trained models.  Thirdly, we hope that the students can state and approach their tasks related to NLP, using the knowledge acquired, conducting experiments, and evaluating the results correctly.

\section{Feedback}

\begin{figure*}[ht]
\centering
\includegraphics[width = .9\textwidth]{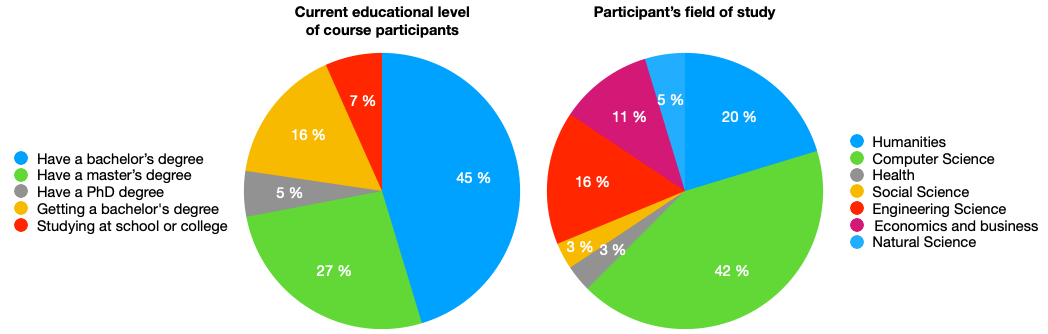}
\caption{The results of survey among course participants. Left:  current educational level. Right: professional area.}
\label{fig:survey}
\end{figure*}

The early feedback we have received so far is positive. Although the course has only been advertised so far to a broader audience, we know that there are two groups interested in the course. First, some students come to study at their own will. Secondly, selected topics were used in offline courses in an inverse classroom format or as additional materials. The students note that our course is a good starting point for studying NLP and helps navigate a broad range of topics and learn the terminology. Some of the students note that it was easy for them to learn in Russian, and now, as they feel more comfortable with the core concepts, they can turn to read detailed and more recent sources. Unfortunately, programming assignments turn out to be our weak spot, as there are challenging to complete, and little feedback on them can be provided.

We ask all participants to fill in a short survey after they enroll in the course. So far, we have received about 100 responses. According to the results, most students (78\%) have previously taken online courses, but only 24\% of them have experience with courses from foreign universities. The average age of course participants is 32 years; most of them already have or are getting a higher education (see Fig.~\ref{fig:survey} for more details). Almost half of the students are occupied in Computer Science area, 20\% have a background in Humanities, followed by Engineering Science (16\%). 

We also ask students about their motivation in the form of a multiple-choice question: almost half of them (46\%) stated that they want to improve their qualification either to improve at their current job (33\%) or to change their occupation (13\%), and 20\% answered they enrolled the course for research and academic purposes. For the vast majority of the student, the reputation of HSE university was the key factor to select this course among other available.

\section{Conclusion}

This paper introduced and described a new massive open online course on Natural Language Processing targeted at Russian-speaking students. This twelve-week course was designed and recorded during 2020 and launched by the end of the year. In the lectures and practical session, we managed to document a paradigm shift caused by the discovery and widespread use of pre-trained Transformer-based language models. We inherited the best of two worlds, showing how to utilize both static word embeddings in a more traditional machine learning setup and contextualized word embeddings in the most recent fashion. The course's theoretical outcome is understanding and knowing core concepts and problem formulations, while the practical outcome covers knowing how to use tools to process text in Russian and English. 

Early feedback we got from the students is positive. As every week was devoted to a new topic, they did not find it difficult to keep being engaged. The ways we introduce the core problem formulations and showcase different tools to process texts in Russian earned approval. What is more, the presented course is used now as supplementary material in a few off-line educational programs to the best of our knowledge. 

Further improvements and adjustments, which could be made for the course, include new home works related to machine translation or mono-lingual sequence-to-sequence tasks and the development of additional materials in written form to support mathematical calculations, avoided in the video lecture for the sake of time.

\bibliography{custom}
\bibliographystyle{acl_natbib}

\end{document}